\documentclass[10pt,twocolumn,letterpaper]{article}

\usepackage{cvpr}
\usepackage{times}
\usepackage{epsfig}
\usepackage{graphicx}
\usepackage{amsmath}
\usepackage{amssymb}
\usepackage{subfigure}
\usepackage{algorithm}
\usepackage{algpseudocode}
\usepackage{bbold}
\usepackage{multirow}
\graphicspath{ {images/} }

\usepackage[breaklinks=true,bookmarks=false]{hyperref}

\cvprfinalcopy 

\def\cvprPaperID{1446} 

\begin{document}

\title{Chinese/English mixed Character Segmentation as Semantic Segmentation}

\author{Huabin Zheng\\
	Sun Yat-sen University\\
	Guangzhou, China\\
	{\tt\small zhhuab@mail2.sysu.edu.cn}
	\and
	Jingyu Wang\\
	iPIN\\
	Shenzhen, China\\
	{\tt\small wangjingyu@ipin.com}
	\and
	Zhengjie Huang\\
	Sun Yat-sen University\\
	Guangzhou, China\\
	{\tt\small hzhengj@mail2.sysu.edu.cn}
	\and
	Yang Yang\\
	iPIN\\
	Shenzhen, China\\
	{\tt\small yangyang@ipin.com}
	\and
	Rong Pan\\
	Sun Yat-sen University\\
	Guangzhou, China\\
	{\tt\small panr@sysu.edu.cn}
}

\maketitle

\begin{abstract}
	OCR character segmentation for multilingual printed documents is difficult due to the diversity of different linguistic characters. Previous approaches mainly focus on monolingual texts and are not suitable for multilingual-lingual cases. In this work, we particularly tackle the Chinese/English mixed case by reframing it as a semantic segmentation problem. We take advantage of the successful architecture called fully convolutional networks (FCN) in the field of semantic segmentation. Given a wide enough receptive field, FCN can utilize the necessary context around a horizontal position to determinate whether this is a splitting point or not. As a deep neural architecture, FCN can automatically learn useful features from raw text line images. Although trained on synthesized samples with simulated random disturbance, our FCN model generalizes well to real-world samples. The experimental results show that our model significantly outperforms the previous methods.
\end{abstract}

\section{Introduction}

\begin{figure*}[!ht]
	\begin{center}
		\includegraphics[width=0.9\linewidth]{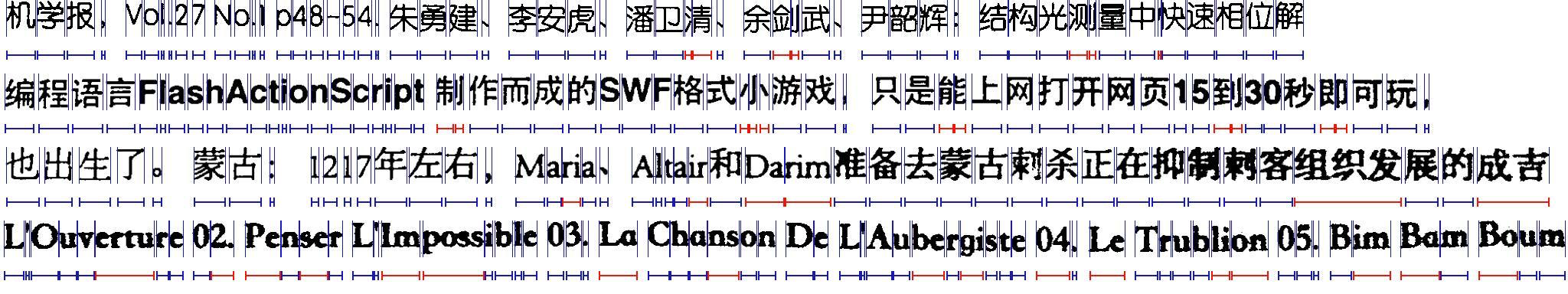}
	\end{center}
	\caption{
		The main challenge of Chinese/English mixed character segmentation arises from the coexistence of disconnected structure in Chinese and touching characters. Correct and wrong segments are colored as blue and red, respectively.
	}
	\label{fig:wrong_segment}
\end{figure*}

Character segmentation plays an important role in optical character recognition (OCR) pipeline~\cite{sahare2016review}. One major reason for poor recognition accuracy in OCR system is the error in character segmentation. Some previous researches~\cite{chen2000segmentation, choudhary2013new, himamunanto2013javanese, hwang1998character, mei2013chinese, nomura2005novel, roy2012multi, senapati2012novel, tse2007ocr} achieve high performance on monolingual texts, but rely on feature engineering specific to single character style. Other researches~\cite{ garain2002segmentation, lee1996new, wang2009high, 1380423} work on multilingual cases but introduce complex processing pipelines. Actually, it's difficult to manually design a set of features suitable for multilingual scene. Thus a mixture of multiple languages presents a challenge for existing character segmentation methods. Chinese/English mixed case is especially difficult due to the coexistence of touching characters and Chinese disconnected structure, as shown in Figure \ref{fig:wrong_segment}. For those ignorant of both languages, it's confusing that a Chinese character with disconnected structure (\eg those in Figure~\ref{fig:wrong_segment}) should \textit{not} be splitted apart, but a pair of touching neighboring English characters (\eg ``DL'' or ``AI") should be splitted apart. Traditional projection based method will falsely break up an intact Chinese character with disconnected structure. A more advanced method~\cite{mei2013chinese} with a connected regions merging phase, tends to falsely take ``DL" as an intact character. In order to correctly perform segmentation, a model should implicitly or explicitly remember all valid characters in both languages. Moreover, in order to deal with various possible font types and sizes, the model should automatically learn necessary features to recognize a valid character since it is cumbersome to specifically design features for every font.

Nowadays we know the ability of deep neural networks to perform automatic feature learning on raw data has significantly advanced the research in various fields of computer vision. Semantic segmentation is among these fields. Fortunately, the problem of multilingual character segmentation can be reframed as the problem of two-class semantic segmentation. To be specific, given a text line image, we classify each horizontal pixel into two categories: splitting point or not. With this problem re-defined, we can utilize those successful deep architectures in recent progress of semantic segmentation. Among them we choose fully convolutional networks (FCN).

In this paper, we reframe OCR character segmentation as semantic segmentation and propose a FCN architecture to solve it. We train our FCN model on synthesized samples with simulated random disturbance and show that it is able to

\begin{enumerate}
	\item significantly outperform previous methods on Chinese/English mixed printed document images;
	\item generalize well from simulated disturbance to real-word disturbance introduced by photographing;
	\item generalize well across different text content styles;
	\item generalize well across different font styles in most cases;
	\item nicely handle disconnected structure and touching characters.
\end{enumerate}

\section{Related Work}

{\bf Previous Approaches}\hspace{1mm} Projection based method is among the simplest approach for OCR segmentation. It calculates the average grey value for each pixel column then split every blank region in the middle, making it vulnerable to disconnected structure and touching characters. Recently, improved methods have been proposed but are only specific for single language~\cite{chen2000segmentation, choudhary2013new, himamunanto2013javanese, hwang1998character, mei2013chinese, nomura2005novel, roy2012multi, senapati2012novel, tse2007ocr}. Other researches exploit complex processing pipelines and hand-crafted rules to tackle multilingual cases~\cite{ garain2002segmentation, lee1996new, wang2009high, 1380423}. There are also researches on handwritten character segmentation~\cite{wshah2009segmentation, tan2012new, kaur2014segmentation, dave2015segmentation}. Compared with printed characters, handwritten characters often require nonlinear splitting paths rather than vertical splitting lines, which is not necessary for regular font types in normal printed documents. 

{\bf Semantic Segmentation}\hspace{1mm} Semantic segmentation is a sub-field of computer vision. Compared with recognition problem, it progresses from coarse to fine inference by making a prediction directly at every pixel~\cite{long2015fully, noh2015learning}. To this end it requires the output size of model to match original input size. However, normal convolutional layer used in recognition only maintains or reduces the size of feature maps, so comes the deconvolution layer.

{\bf Devonlutional Layers} Deconvolution is also called up-convolution~\cite{simo2016learning} or fractionally-strided convolution~\cite{johnson2016perceptual}. It is typically used for expanding the size of feature maps in FCN architecture.

{\bf Fully Convolutional Networks} FCN is prevalent in the research of semantic segmentation and object detection. The key feature that distinguishes FCN from CNN is that it is easy to control the output size of FCN via deconvolution. Therefore, FCN is also widely used in tasks where both input and output are images. For example, Simo-Serra \etal~\cite{simo2016learning} use FCN to simplify sketch drawing.

\begin{figure}[!ht]
	\begin{center}
		\subfigure[Several training samples]{
			\label{fig:several_samples}
			\includegraphics[width=0.95\linewidth]{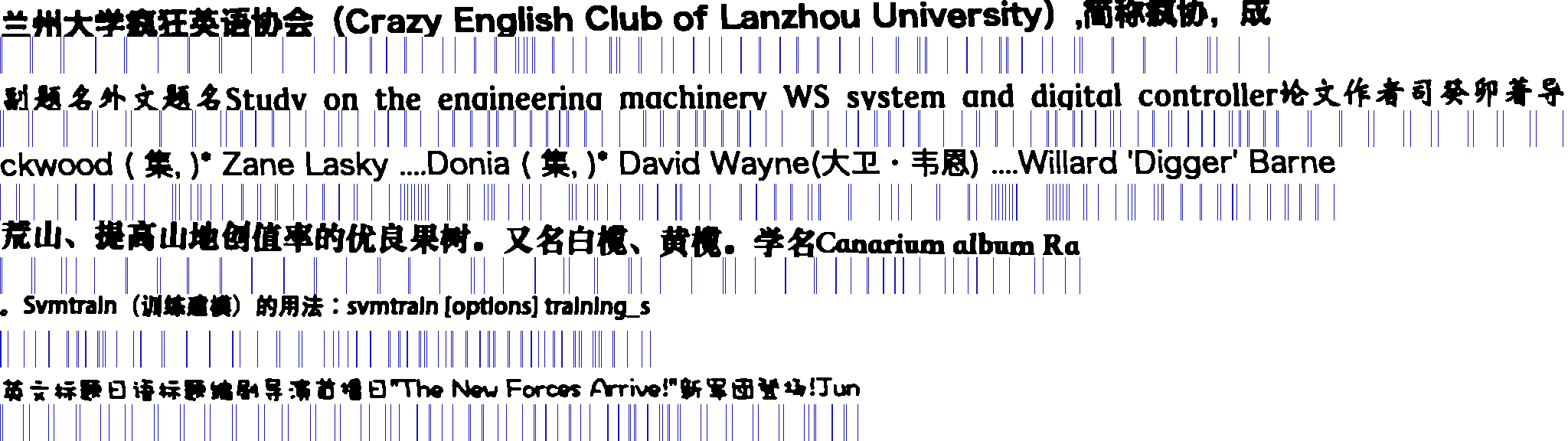}
		}\\
		\subfigure[Zoom in to a region]{
			\label{fig:part_sample}
			\includegraphics[width=0.6\linewidth]{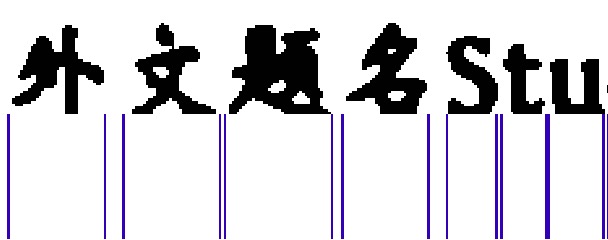}
		}
	\end{center}
	\caption{
		(a) Several training samples and corresponding output ground truths. Notice that the model outputs a vector of length $W$ rather than a matrix of shape $H \times W$. (b) For each character in the image, the left and right margins are given as splitting points. Each splitting point is visualized as a vertical blue lines.
	}
	\label{fig:sample}
\end{figure}

\section{Proposed Approach}

We firstly define the training task in Section~\ref{sec:task_definition}. In Section~\ref{sec:architecture} we propose the detailed architecture of FCN. In Section~\ref{sec:post_processing} we describe the post-processing phase when using the trained model to crop segments. Training data synthesizing process is described in Section ~\ref{sec:training_data_generation}. To deal with imbalanced classes problem, we use a dynamic weighted binary cross entropy loss, which is defined in Section~\ref{sec:dynamic_loss}.

\subsection{Training Task Definition} \label{sec:task_definition} 

In semantic segmentation form, our model is to classify each horizontal pixel position into two classes: splitting point as positive class and non-splitting point as negative class. Formally, given an image of height $H$ and width $W$ as input, a FCN outputs a probability vector $\mathbf{p}$ of length $W$, where
\begin{equation}
\mathbf{p}_i =
\begin{cases}
1 &\parbox[t]{.5\columnwidth}{if the image should be splitted at the $i$-th column;}\\
0 & \text{otherwise.}
\end{cases}
\label{eq:task_define}
\end{equation}

For each character in the image, the left and right margins are given as splitting points. For example, if a character has an extending range from the $m$-th column to the $n$-th column, then we have
\begin{equation}
\begin{aligned}
\mathbf{p}_i &= 1, i = m, n, \\
\mathbf{p}_i &= 0, i = m + 1, m + 2, \dots, n - 1.
\end{aligned}
\end{equation}

See Figure \ref{fig:sample} for several input images and corresponding output ground truths. In this paper we have $H = 48$ and $W = 2048$.

\subsection{Architecture}  \label{sec:architecture}

In a typical FCN architecture, a down-convolution block and an up-convolution block are to reduce and expand the size of feature maps, respectively~\cite{simo2016learning}. Each down-convolution is composed of a convolutional layer, a batch normalization layer~\cite{icml2015_ioffe15}, a max-pooling layer and an activation layer. Each up-convolution is composed of a deconvolutional layer, a batch normalization layer and an activation layer. In research of semantic segmentation, FCN is used to restore both the width and height of original input images~\cite{long2015fully, noh2015learning}.

\begin{figure}[!ht]
	\begin{center}
		\includegraphics[width=1\linewidth]{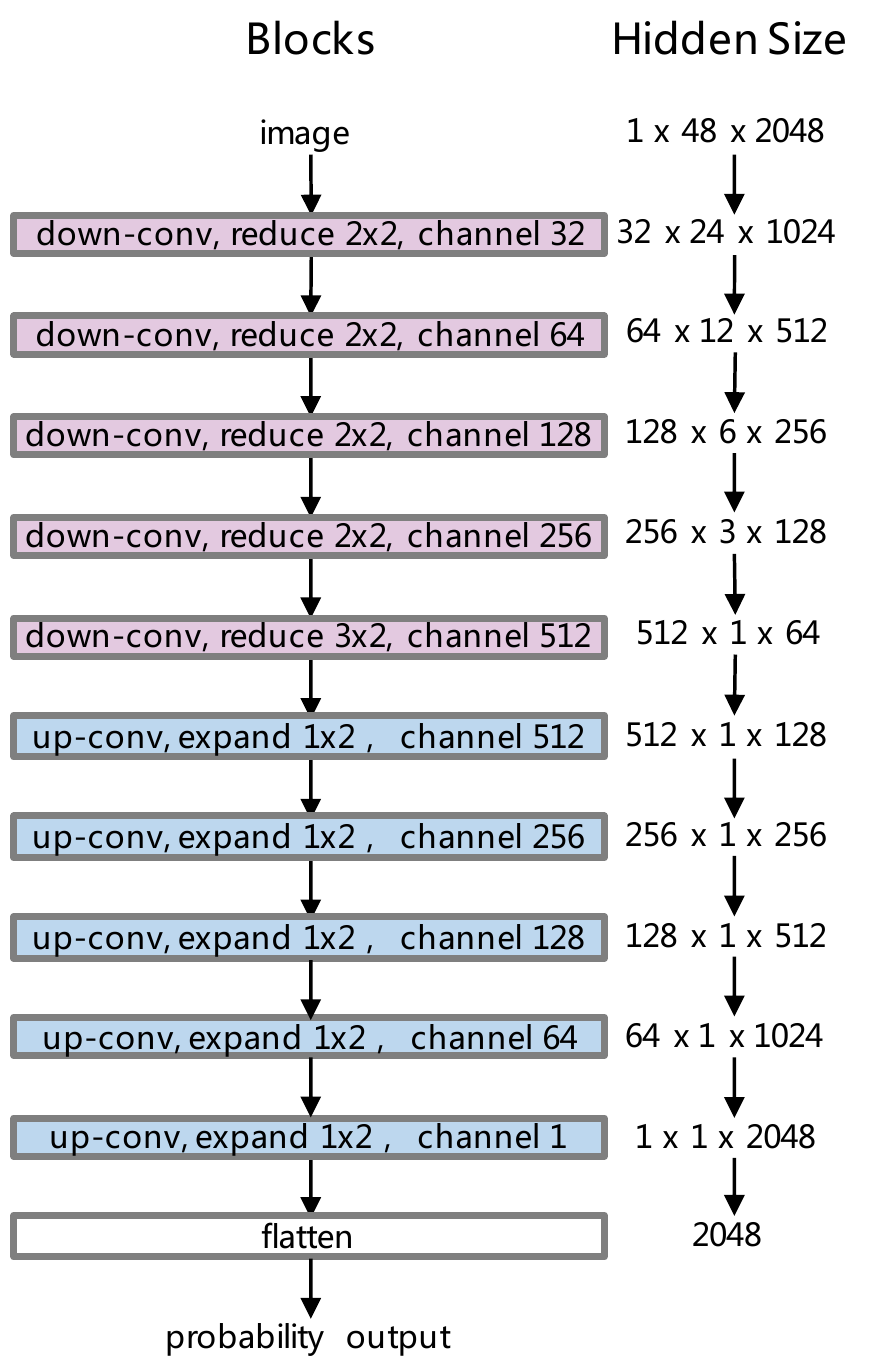}
	\end{center}
	\caption{Our FCN architecture is composed of 5 down-convolution blocks and 5 up-convolution blocks. The activation layer of the last down-convolution block is a sigmoid layer.}
	\label{fig:FCN}
\end{figure}

Our FCN architecture is almost the same as typical ones except that only the width of input images need to be restored. As defined in Section~\ref{sec:task_definition}, Eq.~(\ref{eq:task_define}), it simply outputs a vector of length $W$, which is equivalent to an image of shape $1 \times W$. To this end, deconvolutional layers in our FCN expand the width of feature maps but maintain the height. See Figure~\ref{fig:FCN} for details.

All activation layers except the last one are the ReLU layer~\cite{icml2010_NairH10}. The last one is a sigmoid layer producing a probability output vector.

\begin{figure}[!ht]
	\begin{center}
		\subfigure[Partial input image]{
			\label{fig:post_processing_1}
			\includegraphics[width=0.8\linewidth]{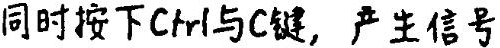}
		}\\
		\subfigure[FCN's output probability]{
			\label{fig:post_processing_2}
			\includegraphics[width=0.8\linewidth]{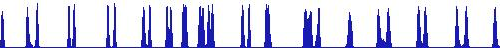}
		}\\
		\subfigure[\textbf{Step 1:} thresholding]{
			\label{fig:post_processing_3}
			\includegraphics[width=0.8\linewidth]{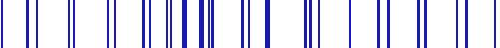}
		}\\
		\subfigure[\textbf{Step 2:} select middle points]{
			\label{fig:post_processing_4}
			\includegraphics[width=0.8\linewidth]{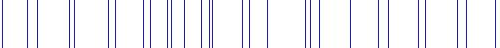}
		}\\
		\subfigure[\textbf{Step 3:} get candidate segments]{
			\label{fig:post_processing_5}
			\includegraphics[width=0.8\linewidth]{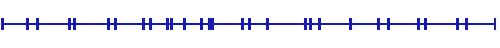}
		}\\
		\subfigure[\textbf{Step 4:} discard blank segments]{
			\label{fig:post_processing_6}
			\includegraphics[width=0.8\linewidth]{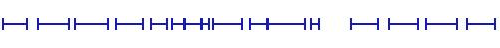}
		}
	\end{center}
	\caption{
		Post-processing steps. (a) Partial input image to FCN. (b) Corresponding part of FCN's output probability vector. (c) \textbf{Step 1:} convert the probability vector into binary vector according to a threshold. (d) \textbf{Step 2:} for each contiguous positive segments, select the center point as splitting point. (e) \textbf{Step 3:} each pair of adjacent splitting points form a candidate segment. (f) \textbf{Step 4:} discard the blank segments and finally output a list of bounding line segments.
	}
	\label{fig:post_processing}
\end{figure}

\subsection{Post-processing for Cropping} \label{sec:post_processing}

As described in Section~\ref{sec:task_definition}, during training phase only two splitting points are given as positive ground truth for each character. During prediction, however, points in adjacent region of a true positive point are usually classified to be positive as well. It's inevitable because neural networks cannot fit the data exactly. In fact, those surrounding points could also be valid splitting points.

Another problem arises when we want to really crop out characters for downstream recognition, because a segment bounded by two adjacent splitting points could contain a character or just the blank between characters. The blank segments should be discarded.

Therefore, we propose a simple post-processing procedure to deal with the issues above. Firstly, the probability vector output $\mathbf{p}$ of FCN is converted to binary vector according to a threshold (\eg 0.5). Secondly, for each contiguous positive segments, the center point is selected as the splitting point. Thirdly, each pair of adjacent splitting points form a candidate segments. Finally, we discard the blank segments and output a list of bounding line segments. Complete pipeline is shown in Figure~\ref{fig:post_processing}.

\subsection{Synthesizing Training Data} \label{sec:training_data_generation}

We focus on camera photographs of Chinese/English mixed printed documents with various fonts. Compared with using scanner, camera photographing introduces much more noises, blurring, rotation and distortion, thus makes the problem more challenging. To approximate these real-world disturbance, we first plot clean texts onto blank image with a size of $2048 \times 48$ then successively apply four simulated random disturbance: rotation, erosion, dilation and Gaussian blurring. Finally we binarize the grey-scale images according to threshold $160$. We keep track of the left and right margins of each character in process above and finally convert them into corresponding binary mask vector as ground truth. Several training input images and output vectors are visualized in Figure \ref{fig:sample}.

\subsection{Dynamic Weighted Binary Cross Entropy} \label{sec:dynamic_loss}

Our model performs binary classification at each of the 2048 horizontal positions. Therefore, given FCN's output probability vector $\mathbf{p}$ and the ground truth vector $\mathbf{q}$, we can simply define the binary cross entropy loss function as
\begin{equation}
\mathcal{L}(\mathbf{p}, \mathbf{q}) = - \sum_{i, \mathbf{q}_i = 1} \log \mathbf{p}_i - \sum_{i, \mathbf{q}_i = 0} \log (1 - \mathbf{p}_i).
\end{equation}

However, the ratio of positive and negative ground truths is not balanced. Most of the ground truths are negative as shown in Figure \ref{fig:sample}. In our experiments unbalanced classes slow down model convergence or even make the model stuck in a local optimum. When stuck, the model predicts negatively at each horizontal position. To tackle this issue, we define a weighted binary cross entropy loss function
\begin{equation} \label{eq:dynamic_loss}
\mathcal{L}(\mathbf{p}, \mathbf{q}) = - \alpha \sum_{i, \mathbf{q}_i = 1} \log \mathbf{p}_i - \beta \sum_{i, \mathbf{q}_i = 0} \log (1 - \mathbf{p}_i),
\end{equation}
where $\alpha + \beta = 1$.

We initialize $\alpha$ to 0.9 and $\beta$ to 0.1. After each iteration we use a heuristic rule to dynamically adjust the weights according to the average positive accuracy $acc_{pos}$ and the average negative accuracy $acc_{neg}$ of the last mini-batch, where
\begin{align}
\label{eq:acc_pos}
acc_{pos} &= \sum_{i, \mathbf{q}_i = 1} \mathbb{1}(\mathbf{p}_i > 0.5) / \sum_{i, \mathbf{q}_i = 1} 1, \\
\label{eq:acc_neg}
acc_{neg} &= \sum_{i, \mathbf{q}_i = 0} \mathbb{1}(\mathbf{p}_i < 0.5) / \sum_{i, \mathbf{q}_i = 0} 1.
\end{align}
If $acc_{pos} < acc_{neg}$ then we increase $\alpha$ and decrease $\beta$, otherwise we increase $\beta$ and decrease $\alpha$. Our strategy can balance the model performance on positive and negative classes throughout training and speed up convergence. See Algorithm \ref{alg:dynamic_loss} for details.

\begin{algorithm}
	\caption{Heuristic Rules for Dynamic Loss}\label{alg:dynamic_loss}
	\begin{algorithmic}[1]
		\State $\alpha \gets 0.9$
		\State $\beta \gets 0.1$
		\While{training}
		\State get mini-batch data for this iteration
		\State update model towards minimizing Eq.~(\ref{eq:dynamic_loss})
		\State compute $acc_{pos}$ according to Eq.~(\ref{eq:acc_pos})
		\State compute $acc_{pos}$ according to Eq.~(\ref{eq:acc_neg})
		\If {$acc_{pos} < acc_{neg}$}
		\State $\delta \gets min(\beta, 0.001)$
		\State $\alpha \gets \alpha + \delta$
		\State $\beta \gets \beta - \delta$
		\Else
		\State $\delta \gets min(\alpha, 0.001)$
		\State $\alpha \gets \alpha - \delta$
		\State $\beta \gets \beta + \delta$
		\EndIf
		\EndWhile
	\end{algorithmic}
\end{algorithm}

\section{Experiments}

\subsection{Datasets}

In this section, we describe the datasets for experiments, including those built by photographing printed documents and those synthesized as described in Section \ref{sec:training_data_generation}.

\subsubsection{Photographed Dataset}

The first dataset is built by photographing as follows. Firstly, text contents are randomly extracted from Baidu Baike corpus, printed with various font types, and photographed with normal phone camera. Secondly, we apply a series of traditional OCR techniques of denoising, binarization, line segmentation \etc to collect a set of text line images. Finally, we hand label the bounding line segment annotations for each character.

\begin{figure}[!ht]
	\begin{center}
		\includegraphics[width=0.95\linewidth]{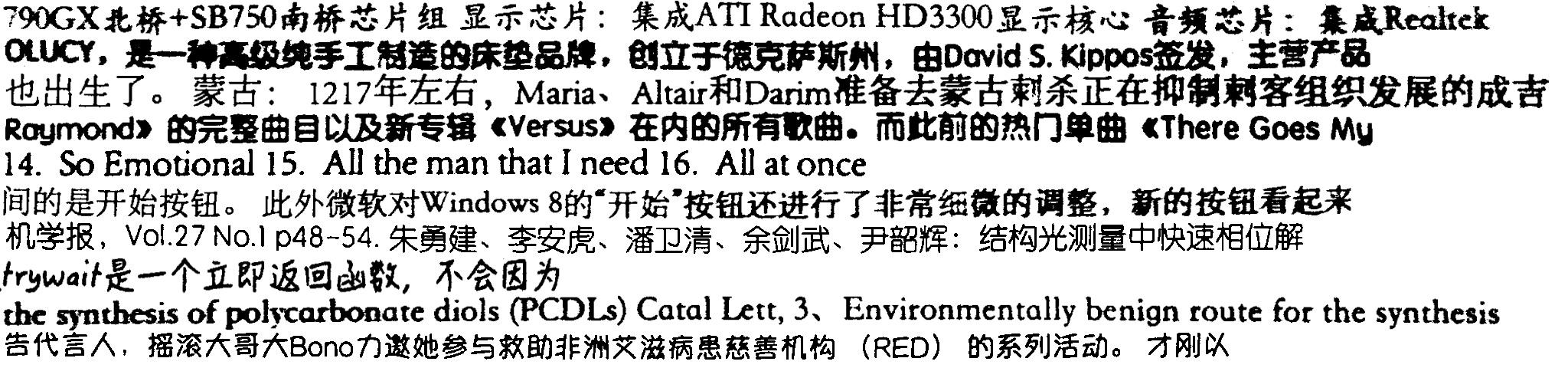}
	\end{center}
	\caption{Several samples of Photo-Normal Dataset.}
	\label{fig:photo_normal}
\end{figure}

Because each text line image sample typically contains several dozens of characters, it takes a long time to annotate even one sample. Thus only 50 text line image samples are finally collected. Nevertheless, they totally contains 2710 characters, which are enough for reliable evaluation. In the following sections, we refer this dataset as ``Photo-Normal". Several samples are shown in Figure \ref{fig:photo_normal}.

\subsubsection{Synthesized Datasets}

The quantity and diversity of Photo-Normal dataset is somewhat limited. Moreover, it can only evaluate our model's generalization ability from simulated disturbance to real-world disturbance. To further evaluate its generalization ability, we synthesize a series of datasets with the combination of two text content styles and 36 font types (specified in next paragraph). Each combination is splitted into training and evaluation parts, producing totally 144 ($2 \times 2 \times 36$) datasets. Among them, one training set contains 3000 samples and one evaluation set contains 30 samples. Total number of characters is more than 10 million.

As for text content styles, the first style called ``normal" simply refers to normal content in Baidu Baike corpus, and the second style called ``chaotic" is acquired by randomly shuffling normal text characters. As for font styles, see Table~\ref{tab:Font_Result} for totally 36 font types used in our experiments.

In the following sections, we refer a dataset of normal text content and font SIMYOU for training as ``Train-Normal-SIMYOU", and so on.

\subsection{Evaluation Metric}

We evaluate segmentation accuracy by matching predictions and ground truths. Given a text line image, FCN and post-processing procedure output a list of $M$ bounding line segments. They are aligned with the ground truth, which is a list of $N$ bounding line segments. For each predicted segment $P_i$ and each true segment $T_j$, we denote the number of $T_j$'s horizontal pixels covered by $P_i$ as $c_{i,j}$ and the number of $T_j$'s horizontal pixels not covered as $u_{i,j}$. Then we say $P_i$ matches $T_j$ if and only if
\begin{align}
min(u_{i,1}, u_{i,2}, \dots, u_{i,N}) & = u_{i,j}, \\
max(c_{i,1}, c_{i,2}, \dots, c_{i,N}) & = c_{i,j}, \\
u_{i,j} & < t_1, \\
c_{i,j} & > t_2, \\
max(c_{i,1}, c_{i,2}, \dots, c_{i,j-1}, c_{i, j+1}, \dots, c_{iN}) & < t_3,
\end{align}
where $t_1, t_2, t_3$ are thresholds.

Taking $t_1 = t_3 = 0$ is equivalent to requiring $P_i$ exactly matched with $T_j$. However, exact matching is not necessary in practice due to some blank space between characters. Therefore we take $t_1=8$, $t_2 = 0$ and $t_3=5$ in our experiments for all the methods we compare, which is fair.

Given the number of matched pairs $K$, we define the segmentation accuracy as

\begin{equation}
acc = \frac{K}{max(M, N)}.
\end{equation}


\subsection{Hyperparameters}

We use mini-batch stochastic gradient descent for 50000 iterations with batch size 8 and momentum coefficient 0.9. For each iteration a mini-batch of samples are randomly selected from training samples. Learning rate is initialized to 0.0001 and divided by 10 at the 20000-th and 40000-th iteration. Training takes approximately 50 minutes on a single GPU (NVIDIA GeForce GTX TITAN X).

\subsection{Quantitative and Qualitative Evaluation}

In this section, we quantitatively evaluate whether our model can generalize
\begin{enumerate}
	\item from simulated disturbance to real-word disturbance,
	\item between normal text content and chaotic text content,
	\item among different font types.
\end{enumerate}

We also qualitatively evaluate its performance on the hardest part of Chinese/English mixed case: coexistence of disconnected components and touching characters.

The generalization over font sizes is trivial and already included throughout these experiments, thus not specifically evaluated.

\subsubsection{Generalization from Simulation to Real-World}

Our FCN is trained on synthesized data because it's difficult to collect a large amount of real word samples with segment annotations. However, simulated disturbance in synthesized samples is definitely not identical to real-world disturbance. FCN must generalize well beyond simulation to deal with photographed printed document images.

To verify this, we train three FCN instances on three datasets and evaluate them on Photo-Normal dataset. Training sets are
\begin{enumerate}
	\item Train-Normal-All: training samples of normal text,
	\item Train-Chaotic-All: training samples of chaotic text,
	\item Train-All-All: union of both above.
\end{enumerate}

Sample counts are 108000, 108000 and 216000, respectively. Each of them contains all the 36 font types. 

In this experiment we compare our approach with four baselines: the traditional projection based method, the region-merging based method designed for Chinese~\cite{mei2013chinese}, the connected component based method designed for English~\cite{choudhary2013new} and Tesseract~\cite{Smith07anoverview}, an open source OCR engine still in active development\footnote{https://github.com/tesseract-ocr/tesseract}. In the following sections they are referred to as PROJ, CN, EN and Tesseract, respectively.

The results are shown in Table \ref{tab:PhotoNormalResult}. Our FCN instances significantly outperform the baselines. Among the first three baselines, PROJ is the simplest model but outperforms CN and EN, because it is not specifically designed for one single language. As a morden OCR engine targeting various languages, Tesseract achieves a decent accuracy but is still outperformed by a large margin. Among the three FCN instances, the one trained on Train-Normal-All achieves the best result, because it has the most similar text content style with Photo-Normal dataset.

All FCN instances achieve over 98\% accuracy. Thus we conclude that they generalize well from simulated disturbance to real-world disturbance.

\begin{table}[!ht]
	\begin{center}
		\begin{tabular}{|c|c|}
			\hline
			Model & Accuracy \\
			\hline
			PROJ & 83.5 \\
			CN~\cite{mei2013chinese} & 72.7 \\
			EN~\cite{choudhary2013new} & 76.0\\
			Tesseract~\cite{Smith07anoverview} & 90.4 \\
			\hline
			Train-All-All & 98.1\\
			Train-Normal-All & \textbf{98.6}\\
			Train-Chaotic-All & 98.2\\
			\hline
		\end{tabular}
	\end{center}
	\caption{Segmentation accuracy on Photo-Normal dataset. ``Train-XXX-All" refers to which dataset our FCN model is trained on. Our model significantly outperforms baselines. The best result is acquired by training on Train-Normal-All, which has the most similar text content style with Photo-Normal dataset.}
	\label{tab:PhotoNormalResult}
\end{table}

\begin{table}[!ht]
	\begin{center}
		\resizebox{0.48\textwidth}{!}{
			\begin{tabular}{|c|c|c|}
				\hline
				\multirow{2}{*}{Training Set} &
				\multicolumn{2}{c|}{Evaluation Set} \\
				\cline{2-3}
				& Eval-Normal-All & Eval-Chaotic-All \\
				\hline
				Train-Normal-All & \textbf{97.8} & 96.0\\
				Train-Chaotic-All & 97.3 & \textbf{97.4} \\
				Train-All-All & 97.5 & 97.2 \\
				\hline
			\end{tabular}
		}
	\end{center}
	\caption{Evaluate generalization ability between normal and chaotic text content styles.}
	\label{tab:Normal_Chaotic_Result}
\end{table}

\subsubsection{Generalization across Text Content Styles}

FCN has the advantage to utilize a wide receptive field on input image to predict at a horizontal pixel. This advantage could be a disadvantage, because it introduces the risk that FCN overfits certain style of text content. For example, if character A is always surrounded by B and C in training text content, FCN may fit such pattern. When it comes to testing text content in which A is surrounded by D end E, FCN will probably make a mistake. We want to know how serious this problem is.

In this section, five datasets are used for training and evaluation:
\begin{enumerate}
	\item Train-Normal-All: training samples of normal text,
	\item Train-Chaotic-All: training samples of chaotic text,
	\item Train-All-All: union of above,
	\item Eval-Normal-All: evaluation samples of normal text,
	\item Eval-Chaotic-All: evaluation samples of chaotic text.
\end{enumerate}

Sample counts are 108000, 108000, 216000, 1080 and 1080, respectively.

The results are shown in Table \ref{tab:Normal_Chaotic_Result}. The best performance on Eval-Normal-All and Eval-Chaotic-All are achieved by training on Train-Normal-All and Train-Chaotic-All, respectively. The first two rows show that our model generalizes well from chaotic style to normal style and only slightly worse from normal style to chaotic style.

This experiment suggests that in practice, if the text content style we want to finally work on can be accessed, it is optimal to train on the same content style. If not, training on chaotic text content still works well.

Note that all of the results in Table \ref{tab:Normal_Chaotic_Result} are worse than results in Table \ref{tab:PhotoNormalResult} because the our simulated disturbance setup is actually more difficult than real-world disturbance.

\subsubsection{Generalization across Font Types} \label{sec:generalize_fonts}

\begin{table}[!ht]
	\begin{center}
		\begin{tabular}{|l|c|c|}
			\hline
			Font & Exclude & All\\
			\hline
			AdobeFangsongStd-Regular  & 98.7 & 99.5 \\
			AdobeHeitiStd-Regular     & 98.6 & 98.6 \\
			AdobeKaitiStd-Regular     & 98.1 & 99.1 \\
			AdobeSongStd-Light        & 99.0 & 99.4 \\
			Baoli                     & 97.5 & 98.1 \\
			Hannotate                 & 98.4 & 98.8 \\
			Hanzipen                  & 95.9 & 96.3 \\
			Hiragino-Sans-GB-W3       & 98.7 & 98.9 \\
			Hiragino-Sans-GB-W6       & 98.9 & 98.3 \\
			Kaiti                     & 98.6 & 98.3 \\
			Lantinghei                & 96.9 & 97.6 \\
			Libian                    & 97.9 & 98.1 \\
			SIMLI                     & 98.8 & 99.3 \\
			SIMYOU                    & 99.2 & 99.3 \\
			\textbf{STCAIYUN}                  & \textbf{51.9} & \textbf{94.7} \\
			STFANGSO                  & 98.0 & 98.2 \\
			STHUPO                    & 93.2 & 96.9 \\
			STHeiti-Light             & 98.3 & 97.7 \\
			STKAITI                   & 98.8 & 99.3 \\
			\textbf{STLITI}                    & \textbf{74.0} & \textbf{83.8} \\
			STSONG                    & 98.2 & 98.8 \\
			STXIHEI                   & 96.4 & 96.7 \\
			\textbf{STXINGKA}                  & \textbf{72.2} & \textbf{91.6} \\
			STXINWEI                  & 98.2 & 98.6 \\
			Songti                    & 98.6 & 98.8 \\
			WawaSC-Regular            & 95.2 & 97.1 \\
			WeibeiSC-Bold             & 98.9 & 98.6 \\
			Yuanti                    & 97.2 & 98.0 \\
			YuppySC-Regular           & 97.6 & 98.2 \\
			msyh                      & 96.8 & 98.7 \\
			msyhbd                    & 97.1 & 98.5 \\
			simfang                   & 99.7 & 99.9 \\
			simhei                    & 98.5 & 99.1 \\
			simkai                    & 98.3 & 98.4 \\
			simsun                    & 98.7 & 98.9 \\
			stzhongs                  & 99.6 & 99.8 \\
			\hline
		\end{tabular}
	\end{center}
	\caption{Evaluate generalization ability across font styles. For each font XXX in the first column, we train a FCN instance on Train-Normal-exclude-XXX dataset and another on Train-Normal-All dataset. Then we evaluate both on corresponding Eval-Normal-only-XXX dataset. Results are shown in the second and third column, respectively. Bad cases are highlighted.}
	\label{tab:Font_Result}
\end{table}

Real-world documents contain various font types. In practice we can include as much font types as possible in training sets to improve generalization. However, some particular fonts of interest may still not be included, which requires the model to generalize across different font types.

To evaluate this generalization ability, three groups of datasets are used in this section:

\begin{enumerate}
	\item Train-Normal-All: training samples of normal text,
	\item Train-Normal-exclude-XXX: training samples of normal text, containing font types except XXX,
	\item Eval-Normal-only-XXX: evaluation samples of normal text, containing only one font type XXX.
\end{enumerate}

Sample counts of each dataset in the three groups are 108000, 105000 and 1080, respectively.

The results are shown in Table \ref{tab:Font_Result}. For each font type, we train FCN on dataset that does not include this font type and dataset that does, corresponding to the second column and third column. The second column shows that FCN generalizes well on unseen font types for the most cases. The third column shows that including corresponding font type during training further improves accuracy.

\begin{figure}[!ht]
	\begin{center}
		\subfigure[STCAIYUN]{
			\label{fig:bad_case_caiyun}
			\includegraphics[width=0.8\linewidth]{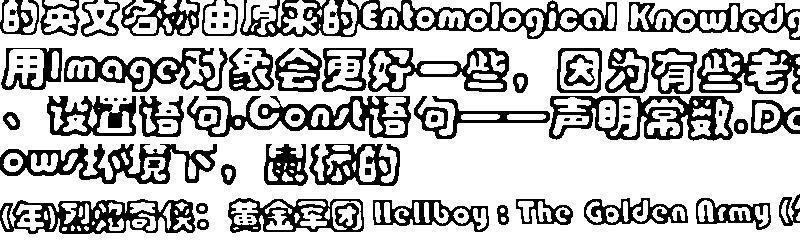}
		}\\
		\subfigure[STLITI]{
			\label{fig:bad_case_lishu}
			\includegraphics[width=0.8\linewidth]{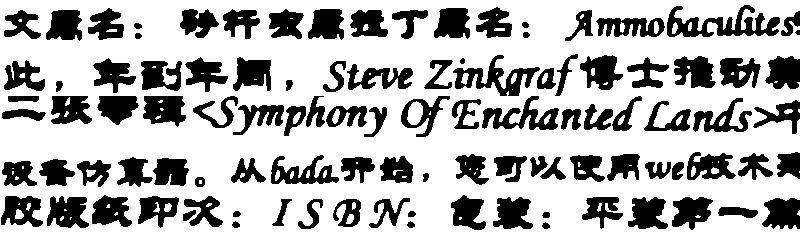}
		}\\
		\subfigure[STXINKA]{
			\label{fig:bad_case_xingkai}
			\includegraphics[width=0.8\linewidth]{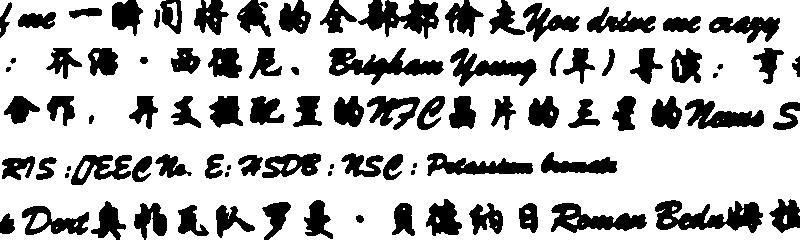}
		}
	\end{center}
	\caption{
		Partial samples of three particular font types on which FCN generalizes badly.
	}
	\label{fig:bad_case}
\end{figure}

Nevertheless, there are several bad cases highlighted in the table: STCAIYUN, STLITI and STXINGKA. Their font styles are illustrated in Figure \ref{fig:bad_case}.

\begin{figure}[!ht]
	\begin{center}
		\subfigure[STLITI]{
			\includegraphics[width=0.8\linewidth]{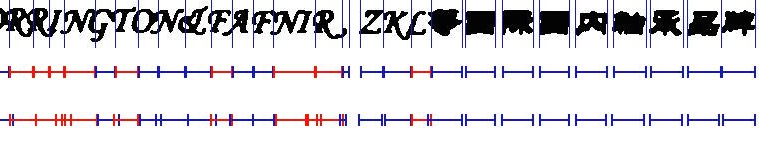}
		}\\
		\subfigure[STXINKA]{
			\includegraphics[width=0.8\linewidth]{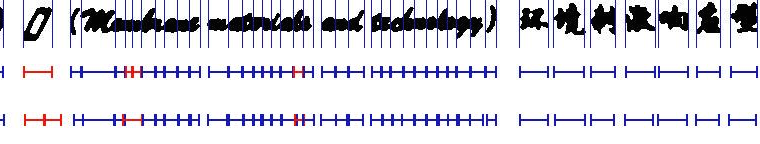}
		}
	\end{center}
	\caption{
		Partial segmentation results on STLITI and STXINGKA. Bounding line segments above and below are predictions and ground truths, respectively. Both FCN instances are trained on Train-Normal-All dataset but still perform illy. Most errors arise from italic English characters.
	}
	\label{fig:bad_case_segment}
\end{figure}

\begin{figure*}[t]
	\begin{center}
		\includegraphics[width=0.9\linewidth]{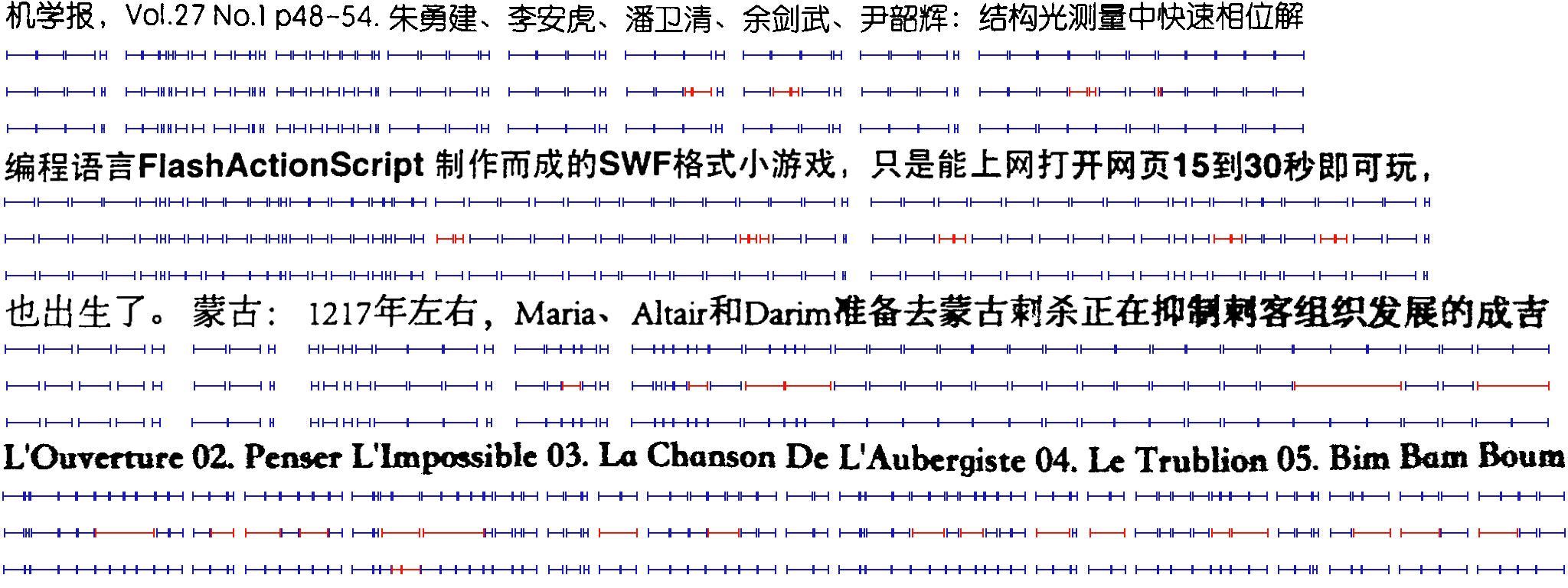}
	\end{center}
	\caption{
		Comparison of projection-based approach and our approach. In each sample there are three rows of bounding line segments, corresponding to ground truths, projection-based method's outputs and our model's outputs, respectively. Correct and wrong bounding line segments are colored as blue and red, respectively. In the last sample FCN falsely splits character ``m" apart probably because FCN recognizes it as two touching ``n". Nevertheless, our FCN handles disconnected components and touching characters well in most cases.
	}
	\label{fig:difficult_cases}
\end{figure*}

The first font type, STCAIYUN, is completely different from others because of its hollow structure. However, when it is included in training set, segmentation accuracy increases from $51.9\%$ back to $94.7\%$. In this case, FCN generalizes badly on particular font type but can restore decent accuracy once it is included again.

However, on the second and third font types, accuracy cannot be restored even when they are included in training set. As shown in Figure \ref{fig:bad_case_segment}, Most segmentation errors arise from English characters. This is because both STLITI and STXINGKA have italic English character style.To properly segment italic characters, the model should predict oblique lines rather than vertical splitting lines, which is impossible in our FCN architecture. 

This experiment shows that our FCN generalizes well on most cases except those with completely different styles or italic styles. 
The first issue can be fixed by including such special font types in training sets. As for the second issue, we will discuss the possible solution in Section \ref{sec:conclusion}.

\subsubsection{Handling Difficult Cases}

The main challenges of Chinese/English mixed character segmentation are two-fold: first, various character widths inside and across languages and second, the coexistence of disconnected structure and touching characters.

Without the first challenge, we can calculate the widths of each connected components in a text line image then take the mode as unified character width, which is used in traditional OCR techniques. Without the second challenge, we can either tune the threshold in projection-based methods to tackle touching characters, or use a region-merging phase~\cite{mei2013chinese} to tackle disconnected structure.

Nevertheless, our FCN architecture handles these difficulties well by automatically utilizing useful features. Typical samples are shown in Figure \ref{fig:difficult_cases}.

\section{Conclusion and Future Work} \label{sec:conclusion}

In this paper we tackle Chinese/English character segmentation for printed document images. By reframing it as a two-class semantic segmentation problem, we take advantage of the successful deep neural architecture called fully convolutional networks (FCN) in the field of semantic segmentation. Trained on synthesized samples with simulated random disturbance, FCN can accurately perform binary classification at each horizontal position on text line images to decide whether this position should be a splitting point or not. Our approach significantly outperforms traditional methods on segmentation accuracy. Experiments show that it is able to generalize from simulated disturbance to real-world disturbance, generalize between normal and chaotic text content styles, generalize among various font types and properly handle the coexistence of disconnected structure and touching characters.

The experimental result in Section \ref{sec:generalize_fonts} shows that our approach performs badly on characters of italic font type because FCN simply predicts vertical splitting lines rather than oblique splitting lines. In addition, there exist even more difficult cases where two characters are so close that they can only be splitted by curved lines. A possible solution and step forward is to reframe character segmentation as an instance segmentation problem. Instance segmentation is also called simultaneous detection and segmentation. In this task, instance-level information and pixel-wise accurate mask for objects are to be estimated~\cite{Liu_2016_CVPR}. Ideally, with instance segmentation every single characters can be curved out exactly and cleanly. In the future we will work on this possible solution.

{\small
	\bibliographystyle{ieee}
	\bibliography{egbib}
}

\end{document}